\pdfoutput=1

\documentclass[11pt]{article}

\usepackage[preprint]{acl}

\usepackage{times}
\usepackage{latexsym}
\usepackage{CJKutf8}
\usepackage{booktabs}
\usepackage{pifont}
\usepackage{makecell}
\usepackage{bm}
\usepackage[english]{babel}
\usepackage{amsmath}
\usepackage{epigraph}
\newcommand{\cmark}{\textcolor{green!50!black}{\ding{51}}}
\newcommand{\xmark}{\textcolor{red!70!black}{\ding{55}}}

\usepackage[T1]{fontenc}

\usepackage[utf8]{inputenc}

\usepackage{microtype}

\usepackage{inconsolata}

\usepackage{graphicx}

\usepackage{tabularx}

\newcommand{\draftcomment}[3]{{\color{#2}[{#1}]$_\text{#3}$}}
\newcommand{\xiang}[1]{\draftcomment{#1}{cyan}{XR}}
\newcommand{\jon}[1]{\draftcomment{#1}{green}{JM}}
\newcommand{\jake}[1]{\draftcomment{#1}{violet}{JB}}
\newcommand{\brihi}[1]{\draftcomment{#1}{magenta}{BJ}}
\newcommand{\hirona}[1]{\draftcomment{#1}{brown}{HA}}
\newcommand{\sheryl}[1]{\draftcomment{#1}{orange}{SM}}
\renewcommand{\brihi}[1]{}
\renewcommand{\jake}[1]{}
\renewcommand{\xiang}[1]{}
\renewcommand{\sheryl}[1]{}
\renewcommand{\hirona}[1]{}
\renewcommand{\jon}[1]{}

\title{Translating the Untranslatable: An Operationalizable Ontology for Untranslatability}

\author{Jacob Bremerman \, Brihi Joshi \, Hirona Arai \, Xiang Ren \, Jonathan May\\
University of Southern California \\
Information Sciences Institute \\
\texttt{\{jbrem,brihijos,hjarai,xiangren,jonmay\}@usc.edu}}

\begin{document}
\maketitle
\begin{abstract}
\textit{Untranslatability}—cases where meaning cannot be directly preserved across languages—is well-studied in linguistics but underexplored in NLP. As machine translation (MT) systems improve on standard benchmarks, their limitations increasingly concentrate in such cases, where translation cannot be reduced to one-to-one equivalence.
We introduce a structured ontology of untranslatability along with a taxonomy of compensation strategies, which are specific techniques to convey meaning under these untranslatable circumstances.
We operationalize this framework into a multilingual dataset of untranslatable sentences paired with strategy-based translations, enabling controlled analysis of translation behavior.
Initial human preference studies suggest that translation quality depends on the strategy used, with consistent preferences for outputs that include explanatory context, known as the Annotation compensation strategy. Our framework and dataset provide a foundation for studying and modeling strategy-informed machine translation.

\end{abstract}

\section{Introduction}
\begin{quote}
\textit{``The limits of my language mean the limits of my world.''}\\
\hfill --- Ludwig Wittgenstein, \textit{Tractatus Logico-Philosophicus} (1922)
\end{quote}
Recent advances in machine translation (MT) and large language models (LLMs) have greatly improved translation quality across many language pairs \cite{Ustun2024AyaMA}, leading some to view translation as a largely `solved problem.'\footnote{A famous article by The Economist claimed Machine Translation to be an `almost solved problem', due to widely available training data. \url{https://www.economist.com/science-and-technology/2024/12/11/machine-translation-is-almost-a-solved-problem}} However, as benchmark performance improves, the remaining challenges increasingly involve long-tail cases where meaning cannot be fully preserved across languages \cite{zhu-etal-2024-multilingual}. These cases are often described as \textit{untranslatability}: situations in which no single translation can capture all aspects of the source meaning because of linguistic, cultural, or stylistic differences \cite{cui-untranslatability}.

\begin{figure}[t!]
    \centering
    \includegraphics[width=\linewidth]{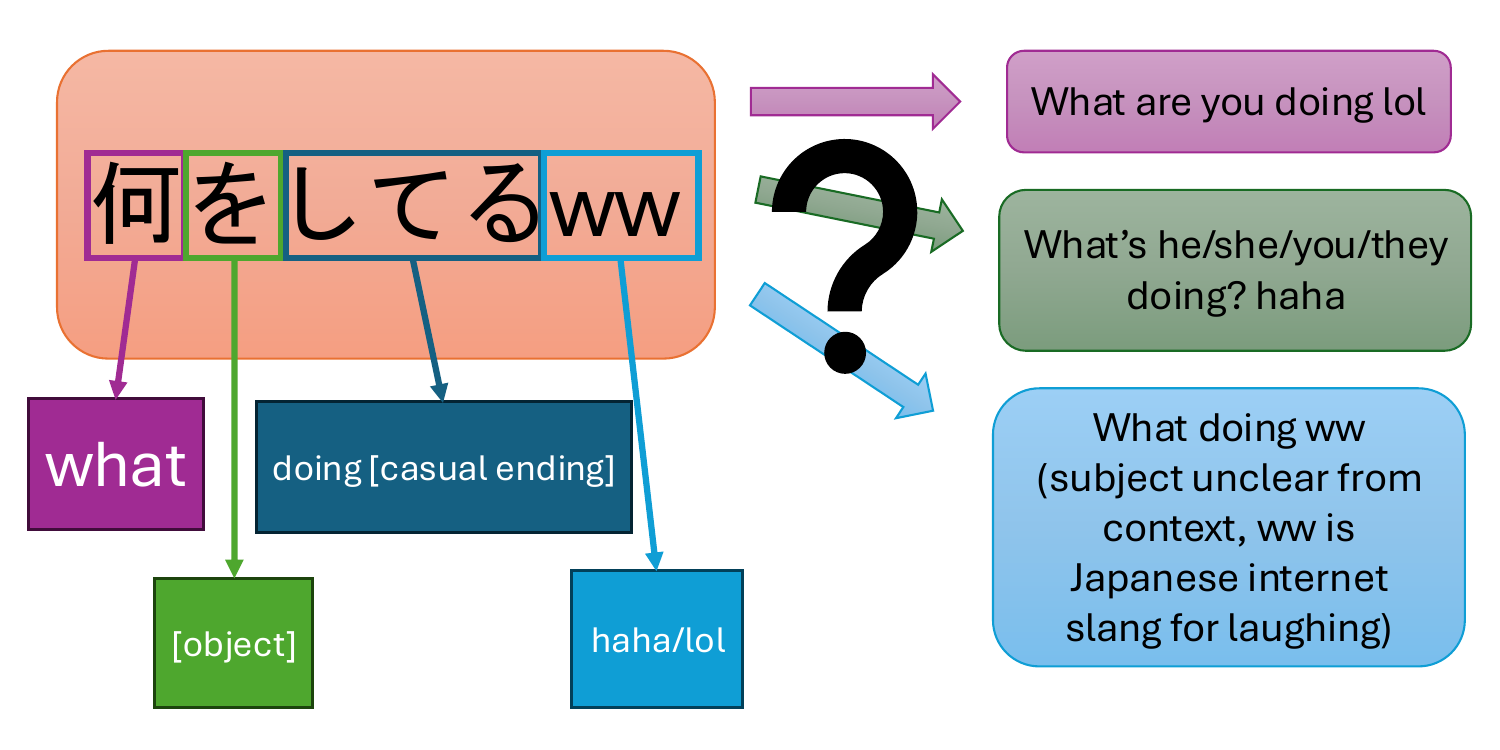}
    \caption{An example of the untranslatability phenomenon.  Due to differences between Japanese and English, the full range of the original meaning has no single translation in English that preserves all meaning.  The appropriate behavior of both human and machine translators is unclear.}
    \label{fig:mot}
\end{figure}

Untranslatability is well established in linguistics and translation studies, arising from mismatches between languages \cite{kitamura2009cultural}. It appears in idioms, slang, wordplay, cultural references, and grammatical distinctions without direct equivalents. As shown in Figure \ref{fig:mot}, even simple utterances may admit multiple valid interpretations that cannot all be preserved in a single translation. Despite this, NLP commonly treats translation as a one-to-one mapping between semantically equivalent texts. In untranslatable cases, however, no single output fully captures the source meaning, requiring decisions about how meaning should be conveyed under linguistic and cultural constraints.

\begin{figure*}[t!]
    \centering
    \includegraphics[width=0.98\linewidth]{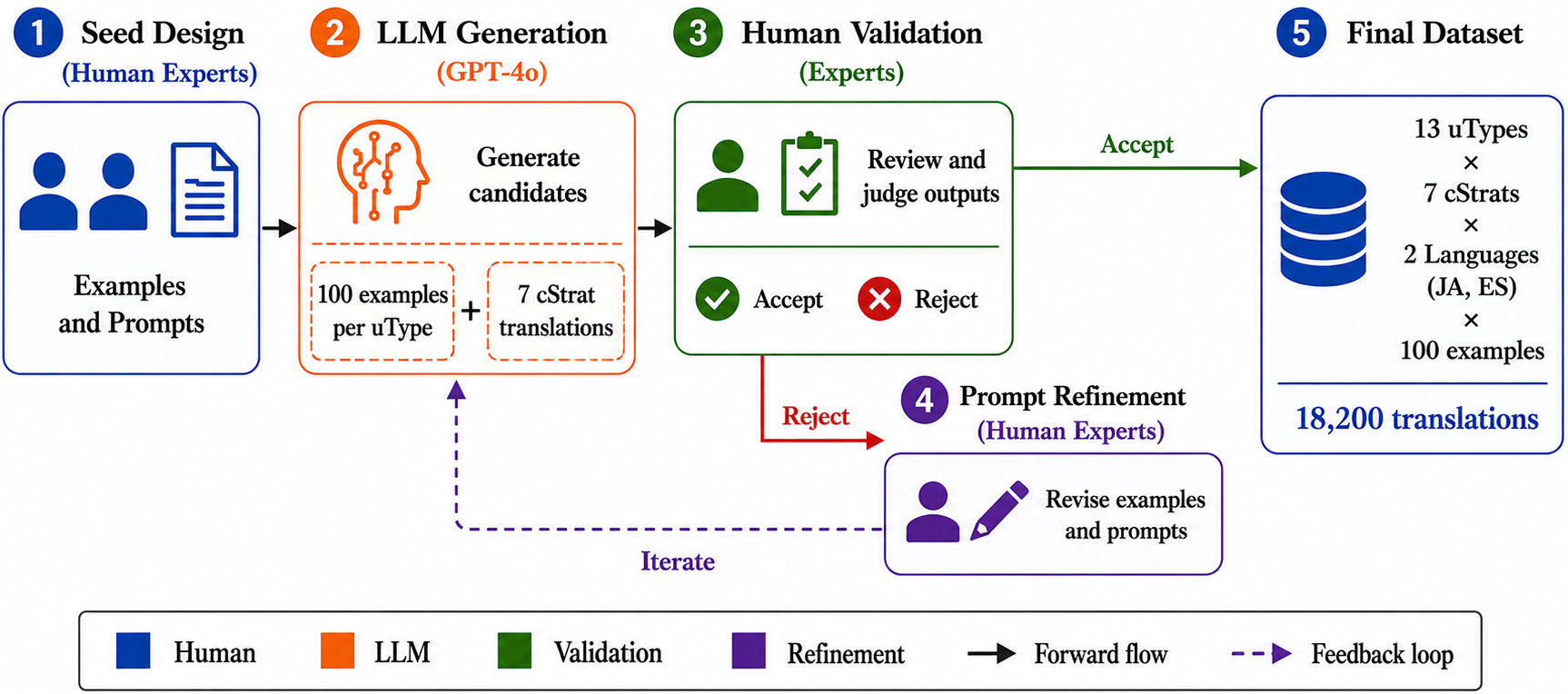}
    \caption{A visualization of the iterative process for generating the untranslatable sentences.  Human experts produce seed examples and prompts, judge outputs and iterate on examples and prompts until outputs are judged to be proper examples of the uType in question.}
    \label{fig:pipe}
\end{figure*}

We argue that untranslatability should be treated as a structured phenomenon in MT rather than a collection of edge cases. To this end, we introduce a framework that categorizes different sources of translation mismatch and the strategies used to convey meaning when direct translation is not possible. We formalize this framework in Section \ref{sec:types}.




Our literature review found no large-scale resource suitable for computational experiments on untranslatability. 
While recent works on \textit{Transcreation}~\citep{khanuja-etal-2024-image} have focus on translation of objects relevant to a particular culture, we believe that unstranslatability is a broader umbrella term, encompassing culture-based transcreation.
To operationalize our ontology, we construct a multilingual dataset of untranslatable sentences paired with translations generated in a target language (here, English) using different compensation strategies. The dataset, created through an iterative process involving human experts and LLMs (Figure \ref{fig:pipe}), enables controlled analysis of translation behavior across different kinds on untranslatability.

Using this dataset, we conduct human preference studies to examine how compensation strategies affect perceived translation quality.
In particular, we ask how preferences vary across different types of untranslatability. 
Our findings suggest that translation quality depends not only on fidelity to the source text, but also on the suitability of the strategy used. 
We consistently observe preferences for translations that include additional explanatory context, highlighting a gap between standard MT outputs and human expectations.

This work establishes a foundation for studying untranslatability in NLP, where no scalable framework currently exists. 
Our contributions are: (1) a structured ontology of untranslatability for MT; (2) a taxonomy of compensation strategies; (3) a multilingual dataset operationalizing this framework; (4) initial empirical evidence on strategy-dependent translation preferences; and (5) a formulation of strategy-informed machine translation.\footnote{Code available at \url{https://github.com/jlbrem/untranslatability}. Data available at \url{https://huggingface.co/collections/INK-USC/untranslatability}.}

\section{Related Works}\label{sec:rel}

While untranslatability has been extensively studied in theoretical linguistics and translation studies, its treatment in NLP remains fragmented. Prior work in NLP typically focuses on isolated phenomena such as idioms, slang, or honorifics, without a unified framework for reasoning about translation under semantic mismatch. In this section, we situate our work within these lines of research and highlight the absence of a structured ontology connecting them.

\subsection{Untranslatability}

\paragraph{Theoretical Translation.}
Untranslatability has long been explored in theoretical fields of language study such as linguistics and translation studies. \citet{cui-untranslatability} offers definitions of untranslatability and compensation strategy along with categorizations of their different types. Other works ~\citep{kitamura2009cultural, puchala-untranslatability,riabovol2023problem} provide further interpretation and examples of untranslatability across languages. These studies are valuable for understanding the phenomenon conceptually, but they typically provide limited examples and do not offer a formal structure that can be directly operationalized in NLP systems. A central goal of our work is to bridge this gap by introducing a structured ontology and instantiating it at scale (see. Table~\ref{tab:dataset-comparison}).

\begin{table*}[t]
\centering
\small
\begin{tabular}{l
>{\centering\arraybackslash}p{1.4cm}
>{\centering\arraybackslash}p{1.4cm}
>{\centering\arraybackslash}p{1.6cm}
>{\centering\arraybackslash}p{2.0cm}
>{\centering\arraybackslash}p{1.6cm}
>{\centering\arraybackslash}p{2.0cm}}
\toprule
\textbf{Dataset} 
& \makecell{\textbf{uTypes} \\ \textbf{classified}} 
& \makecell{\textbf{cStrats} \\ \textbf{classified}} 
& \makecell{\textbf{Several} \\ \textbf{translations}}
& \makecell{\textbf{Downloadable} \\ \textbf{data table}}
& \makecell{\textbf{Human} \\ \textbf{preference}} 
& \makecell{\textbf{Total} \\ \textbf{examples}} \\
\midrule

Ours 
& \cmark 
& \cmark 
& \cmark 
& \cmark 
& \cmark
& \textbf{18,200} \\

\citet{cui-untranslatability}
& \cmark 
& \cmark 
& \xmark
& \xmark 
& \xmark
& $\sim10$ \\

\citet{puchala-untranslatability}
& \cmark 
& \xmark 
& \xmark
& \xmark 
& \xmark
& $\sim10$ \\

\citet{riabovol2023problem}
& \cmark 
& \xmark 
& \xmark
& \xmark 
& \xmark
& $\sim10$ \\

\citet{kitamura2009cultural}
& \xmark 
& \xmark 
& \xmark
& \xmark 
& \xmark
& $\sim10$ \\

\bottomrule
\end{tabular}
\caption{Comparison of our dataset with existing resources.  We are the first downloadable dataset with a high order of example sentences.  Other publications have datasets that would require additional parsing and do not approach the order of magnitude of ours.}
\label{tab:dataset-comparison}
\end{table*}

\paragraph{Machine Translation.}
Untranslatability has also been studied in the context of machine translation, both directly and indirectly. Direct approaches \citep[e.g.,][]{cheng-etal-2014-detecting} focus on detecting untranslatable expressions in specific languages. More commonly, prior work examines particular manifestations of untranslatability without framing them as such. Examples include studies on the translation of poetry \cite{ghazvininejad-etal-2018-neural}, idioms \cite{fadaee-etal-2018-examining} and honorifics \cite{sennrich-etal-2016-controlling}.  While these works provide useful techniques for handling specific phenomena, they address individual aspects of untranslatability in isolation and do not provide a unified framework for reasoning across them.

\subsection{Large Language Models}

\paragraph{LLM Translation.}
Although many LLMs are primarily designed with English in mind, they have demonstrated strong multilingual capabilities, including in machine translation. \citet{zhu-etal-2024-multilingual} show that models such as GPT-4 can perform comparably to translation-specific systems like Google Translate in relatively straightforward settings. Our work complements these findings by focusing on more challenging cases, investigating how LLMs handle instances of untranslatability where direct translation is insufficient.

\paragraph{LLM Data Generation.}
Another relevant line of research explores the use of LLMs for synthetic data generation. \citet{abdullin-etal-2023-synthetic} demonstrate that GPT-4~\citep{openai2024gpt4technicalreport} can be used to synthesize datasets that are useful for training downstream models.
We similarly leverage LLMs to generate data, but with a different objective: to operationalize a structured framework of untranslatability at scale. 
While LLMs enable efficient data generation, they do not address the underlying question of how untranslatability should be systematically defined or represented.

\paragraph{Summary.}
Taken together, prior work either provides theoretical insights into untranslatability without large-scale operationalization, or empirical approaches that address specific phenomena without a unifying framework. 
Our work bridges this gap by introducing a structured ontology of untranslatability and instantiating it in a dataset that enables systematic study within NLP.
\section{A Framework for Untranslatability}\label{sec:types}

To study untranslatability systematically in the context of NLP, it is necessary to move beyond informal descriptions and establish a structured representation of the phenomenon. In this section, we introduce a framework that decomposes untranslatability into two key components: (1) the underlying source of the mismatch between languages, and (2) the strategy used to convey meaning in the presence of that mismatch.




\begin{figure}
    \centering
    \includegraphics[width=1\linewidth]{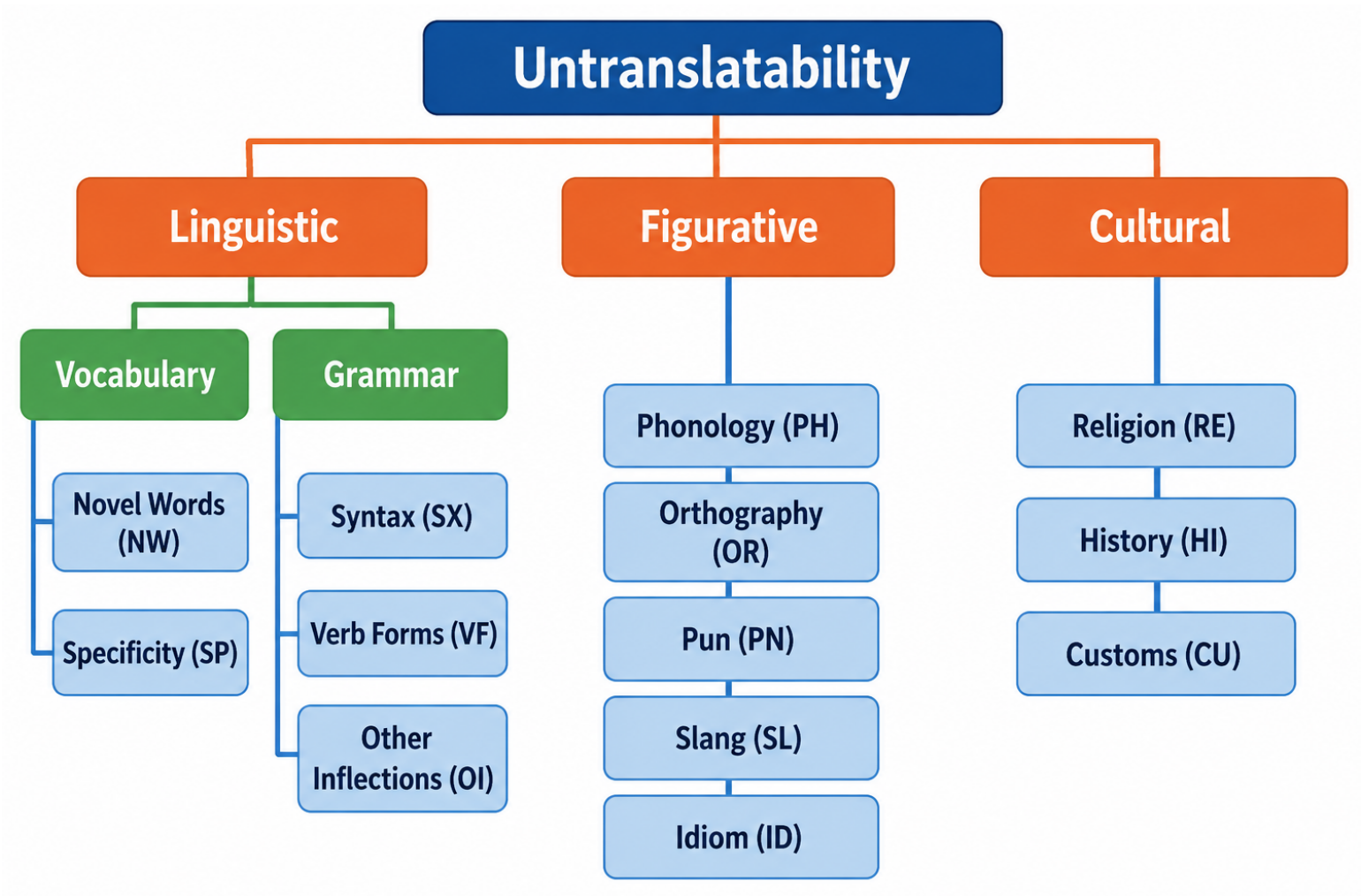}
    \caption{Ontology of uTypes: Refer to Section \ref{sec:types} for more details on definitions and examples.}
    \label{fig:ont}
\end{figure}

\subsection{An Ontology of Untranslatability}
\label{sec:ctype}

We organize untranslatability into a structured ontology of \textit{untranslatability types} (uTypes), which capture the underlying source of mismatch between languages. We group these into three domains: Linguistic, Figurative, and Cultural (Figure~\ref{fig:ont}).

Table~\ref{tab:utypes} provides representative examples of each uType. Full definitions and additional examples are provided in Appendix \ref{sec:appendix-utype}. While constructing the ontology requires human linguistic expertise, it is grounded in prior cross-linguistic work to be extensible to additional languages. This offsets the expense of human experts when expanding the dataset and/or ontology.

\begin{table*}[t]
\centering
\small
\begin{tabular}{p{3cm} p{6cm} p{6cm}}
\toprule
\textbf{Type} & \textbf{Example} & \textbf{Why untranslatable} \\
\midrule
Novel Words (NW) &
\textit{Disfrutamos la sobremesa.} $\rightarrow$ ``We enjoyed the sobremesa.'' &
``Sobremesa'' has no single English word; it refers to time spent talking after a meal. \\
\hline
Specificity (SP) &
\begin{CJK*}{UTF8}{gbsn}叔叔来了。\end{CJK*} $\rightarrow$ ``My uncle is here.'' &
``Uncle'' underspecified in English vs Chinese. \\
\hline
Syntax (SX) &
\begin{CJK*}{UTF8}{gbsn}这个电影我很喜欢。\end{CJK*}  $\rightarrow$ ``This movie I like.''&
Flexible Chinese word order nuance. \\
\hline
Verb Forms (VF) &
\begin{CJK*}{UTF8}{ipxm}行きますか？うん、行く。\end{CJK*} $\rightarrow$ ``Are you going (formal)? Yes, I'll go (informal). &
Formal and informal verb forms carry meaning that English cannot preserve directly. \\
\hline
Other Inflections (OI) &
\textit{Los profesores y las profesoras} $\rightarrow$ ``The male teachers and the female teachers''&
Grammatical gender is explicit in Spanish but not in English. \\
\hline
Phonetics (PH) &
``Peter Piper picked a peck...'' &
Sound patterns such as alliteration cannot usually be preserved alongside meaning. \\
\hline
Orthography (OR) &
\begin{CJK*}{UTF8}{ipxm}wwww\end{CJK*} $\rightarrow$ \begin{CJK*}{UTF8}{gbsn}草\end{CJK*} (www looks like grass and also means lol in Japanese)&
Visual form contributes meaning that does not transfer across scripts. \\
\hline
Pun (PN) &
\textit{El pan está blando.} Sounds like the bread ``is bland'' and ``is talking.''&
Wordplay depends on multiple meanings that are hard to preserve simultaneously. \\
\hline
Slang (SL) &
\begin{CJK*}{UTF8}{gbsn}他很牛逼。\end{CJK*} $\rightarrow$ ``He’s badass.'' &
Meaning is cultural; can't map word-to-word. \\
\hline
Idiom (ID) &
\begin{CJK*}{UTF8}{ipxm}猿も木から落ちる。\end{CJK*} $\rightarrow$ even monkey falls from tree.&
The literal meaning differs from the intended figurative meaning. \\
\hline
Religion (RE) &
\textit{Me crucificaron} $\rightarrow$ so I was crucified.&
Reference depends on shared religion. \\
\hline
History (HI) &
\begin{CJK*}{UTF8}{ipxm}奈良時代\end{CJK*} $\rightarrow$ Nara period&
Historical references require cultural context.\\
\hline
Customs (CU) &
\textit{Le di dos besos.} $\rightarrow$ I gave her two kisses. &
Greeting customs differ across cultures. \\
\bottomrule
\end{tabular}
\caption{Untranslatability types (uTypes) with representative examples.}
\label{tab:utypes}
\end{table*}

\subsection{Compensation Strategies}
\label{sec:cstrat}

When direct translation is not possible, translators employ alternative methods to convey meaning. 
We refer to these as \textit{compensation strategies} (cStrats), which determine how information lost in translation is represented in the target language.

We adapt a categorization from prior work \cite{cui-untranslatability}, defining six primary cStrats. Table~\ref{tab:cstrats} shows representative examples (see Appendix \ref{sec:appendix-utype}).

\begin{table*}[t]
\centering
\small
\begin{tabular}{p{2.2cm} p{5.85cm} p{6.5cm}}
\toprule
\textbf{Strategy} & \textbf{Example} & \textbf{How it works} \\
\midrule
Adaptation (AD) &
Replace a Japanese tongue-twister with an English one &
Preserves the effect, such as humor or rhythm, rather than the literal wording. \\
\hline
Annotation (AN) &
``We enjoyed the sobremesa (post-meal chat typical in Spain)'' &
Adds explanatory information to preserve meaning. \\
\hline
Borrowing (BO) &
``The room has very nice feng-shui.'' &
Keeps the source-language term in the translation. \\
\hline
Calque (CA) &
Literal translation (``long time no see'') &
Preserves  structure and lets reader infer  meaning. \\
\hline
Options (OP) &
Se fue pa casa. $\rightarrow$ ``He/she went home.'' &
Multiple  options when source underspecified. \\
\hline
Paraphrase (PA) &
Iba pedo. $\rightarrow$ ``He was very drunk.'' &
Re-express in plain target-language wording. \\
\bottomrule
\end{tabular}
\caption{Compensation strategies (cStrats) with examples.}
\label{tab:cstrats}
\end{table*}

The most appropriate cStrat for an input depends, in general, on both the uType and the translation context. This suggests that translation involves not only generating text, but also selecting how meaning should be conveyed.

\subsection{Implications for Machine Translation}

Standard MT formulations model translation as token-level prediction:
\[
P(e|f) = \prod_i P(t_i \mid t_1, \dots, t_{i-1}, f),
\]
where $f$ is the source sentence and $e$ the target sentence.

However, untranslatability violates the assumption that a single target sentence can fully preserve source meaning. We instead view translation as a structured process involving identification of the mismatch and selection of a compensation strategy, decomposing $P(e|f)$ as
\[
P(e \mid f, u, m, c)\cdot P(c \mid f, u, m)\cdot P(u \mid f),
\]
where $u$ is the untranslatability type, $m$ the translation medium/context, and $c$ the compensation strategy. This decomposition separates translation into three components: identifying the uType present, selecting how meaning should be conveyed based on the context and uType, and finally generating the actual tokens to be used in a specific language to accomplish this.

This formulation is conceptual rather than directly implemented, but illustrates how untranslatability and strategy selection may be incorporated into MT systems. In practice, such a framework could be integrated into existing MT systems by augmenting decoding with strategy selection or by post-processing outputs for detected instances of untranslatability.

\section{Dataset Construction}\label{sec:data}

To operationalize our framework, we construct a dataset of untranslatable sentences paired with translations that reflect different cStrats. The dataset is designed to enable controlled analysis of translation behavior across different uTypes and cStrats.


\begin{table*}[]
    \centering
    \begin{tabular}{|l|p{4.25in}|}
    \hline
     id	& es-cu-0 \\ 
     \hline
     sentence	&Disfrutaron de una sobremesa larga. \\ 
     \hline
     AD (Adaptation)	& They enjoyed a long after-dinner chat. \\
     \hline
     AN	(Annotation) & They enjoyed a long sobremesa. (In Spanish culture, sobremesa refers to the time spent talking at the table after a meal.) \\
     \hline
     BO	(Borrowing) & They enjoyed a long sobremesa \\
    \hline
     CA	(Calque) & They enjoyed a long over-table. \\
     \hline
     OP (Options) & They enjoyed a long {after-dinner conversation/afternoon/ meal time}. \\
     \hline
     PA (Paraphrase) & They enjoyed a long meal discussion.	\\
     \hline
     NO (Default) &  They enjoyed a long after-dinner conversation. \\
     \hline
     
    \end{tabular}
    \caption{This table includes the headers and one example row from our Spanish split of the dataset (transposed for space).  The id row displays the two letter language code followed by the two letter uType code followed by the index for the example between 0 and 99.  The capital two-letter headers represent the 6 cStrat translations with the `NO' column representing the seventh `default' translation, which uses `no' explicit teaching of cStrats in the prompting.}
    \label{tab:example}
\end{table*}

\subsection{Overview}

Our dataset consists of source sentences exhibiting instances of untranslatability, along with multiple corresponding translations in English. Each source sentence is labeled with its uType, and each translation is generated according to a specific cStrat, as defined in Section \ref{sec:types}.

We focus on Spanish and Japanese as source languages, paired with English as the target language. We choose these languages because of the authors' familiarity as well as access to bilingual speakers for annotations.  For each of the 13 uTypes, we generate 100 examples per language, resulting in 1,300 source sentences per language. For each sentence, we produce translations using six cStrats as well as a default, unprompted translation. This yields a total of 18{,}200 translations across both languages. An example of the dataset structure is shown in Table \ref{tab:example}.

\subsection{Data Generation}

We generate the dataset through an iterative process combining human expertise and LLM-based generation (Figure \ref{fig:pipe}). For each uType and language, human experts first provide a small set of seed examples that illustrate the phenomenon. These examples are then used to prompt an LLM to generate additional sentences exhibiting the same type of untranslatability. 

A similar process is used to generate translations for each compensation strategy. Seed examples for each strategy are used to guide the model in producing translations that reflect the intended behavior. We also include a default generation setting without explicit strategy guidance, representing standard translation behavior.

While LLM-based generation enables large-scale data creation, it introduces potential biases and inconsistencies. We mitigate these issues through iterative prompt refinement and human oversight during generation.

\subsection{Data Validation}

To assess the quality of the generated data, we conduct a human verification study. Language experts review a sample of generated sentences and evaluate whether they correctly represent the intended untranslatability type.

Across sampled examples, approximately 96\% of Spanish and 95\% of English examples are judged to be valid instances of their assigned uType. We observe some variation in quality across types, with certain phenomena being more difficult for the model to generate consistently. Full details of the validation procedure and results are provided in Appendix \ref{sec:details}.

Although examples within each uType share structural characteristics, we observe substantial lexical diversity across generated instances, reflecting variation beyond the seed examples.

Overall, this dataset enables controlled analysis of translation under semantic mismatch by explicitly labeling both untranslatability types and compensation strategies, which goes beyond standard MT evaluation.



\section{Human Preference Study}\label{sec:pref}

With the dataset constructed, we conduct a human preference study to examine how different compensation strategies affect translation quality. Our goal is to obtain initial evidence on how preferences vary across instances of untranslatability.

A key motivation for this study is that different cStrats may be more or less appropriate depending on both the type of untranslatability and the translation context. For example, Borrowing may be suitable for culturally specific terms (e.g., Novel Words), but less appropriate for grammatical mismatches (e.g., Verb Forms). Similarly, strategies that introduce additional explanation, such as Annotation, may be effective in settings where comprehension is prioritized, but less appropriate in contexts requiring brevity.

Based on these observations, we investigate the following research questions: (1) How does the uType of a source sentence affect which cStrat is preferred? (2) How does translation context affect strategy preference? (3) How does the source language influence preferred strategies?

\subsection{Experimental Setup}\label{sec:exp}

We recruit bilingual annotators using Prolific,\footnote{https://www.prolific.com/} selecting participants fluent in English and either Spanish or Japanese. Annotators are not informed about uTypes or cStrats, ensuring that judgments reflect natural preferences.

For each source sentence, annotators are shown seven candidate translations: one for each compensation strategy and one default translation without explicit strategy guidance. Translations are presented in random order, and annotators are asked to rank them based on overall quality and provide brief explanations. The annotation interface is shown in Figure~\ref{fig:ui}.

To analyze variation across uTypes, we sample 50 examples per uType and collect rankings from three annotators per example, resulting in 1,950 total annotations.

We also examine the effect of translation context by modifying the evaluation setup. In this condition, annotators are asked to imagine a specific usage scenario before ranking translations. The interface for this setting is shown in Figure~\ref{fig:context-ui}. We consider two contexts: \textit{textbook}, where the goal is accurate understanding, and \textit{movie}, where the goal is natural and engaging translation.

For the context-based evaluation, we sample 130 examples (10 per uType) and collect annotations from three annotators per example for each context, resulting in 780 additional annotations. We measure inter-annotator agreement using Kendall’s coefficient of concordance (W)~\citep{kendallcoeff}, obtaining 0.58 for Spanish and 0.62 for Japanese, indicating moderate agreement.
\section{Results}

\subsection{Strategy preference varies with untranslatability type}

\begin{figure}
    \centering
    \includegraphics[width=\linewidth]{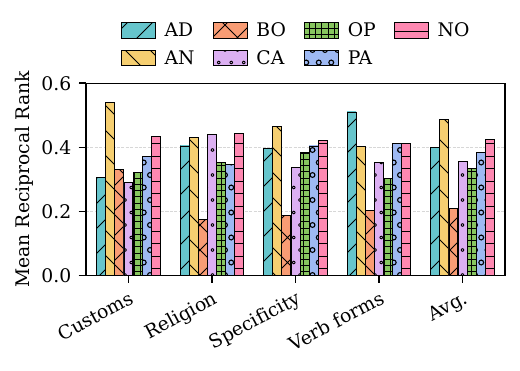}
    \caption{Mean Reciprocal Rank for cStrats based on uType of source sentence. Full results are shown in Figure \ref{fig:gen-es} in the Appendix.  Annotation is most preferred overall. The uTypes RE and VF showing preference for CA and AD, respectively.}
    \label{fig:es-zoom}
    \vspace{-0.3cm}
\end{figure}

We first examine overall trends in the general (context-free) setting. Full results across all uTypes are provided in Appendix Figure \ref{fig:gen-es}, while Figure \ref{fig:es-zoom} highlights representative cases.  We use Mean Reciprocal Rank (MRR), which averages the inverse rank of each strategy across annotators.\footnote{https://en.wikipedia.org/wiki/Mean\_reciprocal\_rank}

Across uTypes, we observe consistent differences in preference between strategies. On average, Annotation (AN) is the most preferred strategy. This is notable because Annotation produces outputs that differ substantially from standard MT behavior by including additional explanatory content. The preference for such outputs suggests that, in cases of untranslatability, users value clarity and completeness over strict adherence to conventional translation formats.

Preferences also vary systematically across uTypes. For example, the Options (OP) strategy performs best for Specificity (SP), where multiple interpretations exist in the target language. In such cases, explicitly presenting alternatives aligns well with the underlying ambiguity.

The Adaptation (AD) strategy exhibits high variability across uTypes. It performs relatively well for linguistic phenomena (e.g., NW, OI, VF), where meaning can often be re-expressed within the target language, but performs worse for figurative and cultural phenomena (e.g., PH, PN, OR, CU, HI), where altering content distorts meaning. These trends are consistent with the intuition that adaptation is more effective when the mismatch is structural rather than cultural.

Borrowing (BO) is least preferred on average, but performs better for culturally grounded cases (CU), where retaining the original term preserves meaning more effectively than translating it. Conversely, for the Religion (RE) uType, Calque (CA) appears more competitive, reflecting shared linguistic structure across related cultural contexts. However, these patterns also depend on the specific language pairs considered.

\subsection{Translation context influences strategy preference}

\begin{figure}
    \centering
    \includegraphics[width=\linewidth]{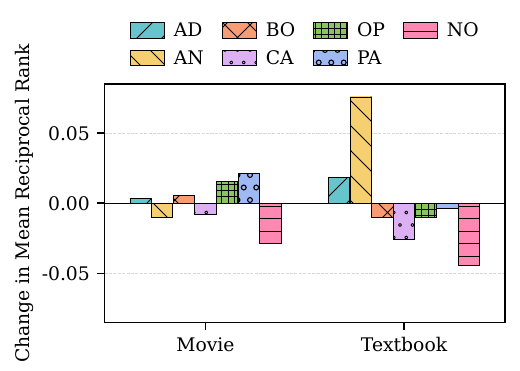}
    \caption{Change in MRR for each compensation strategy based on translation context.  Annotation is much more preferred for Textbook and less preferred for Movie contexts.}
    \label{fig:mmr-delta}
    \vspace{-0.15cm}
\end{figure}

We next examine how preferences vary across translation contexts. Full results are shown in Appendix Figure \ref{fig:mmr-context}, while Figure \ref{fig:mmr-delta} highlights differences relative to the general setting.

We observe that the default translation (no strategy prompting) is consistently less preferred in both contextual settings. This suggests that explicitly applying compensation strategies better aligns translations with user expectations in context-dependent scenarios.

Preferences for specific strategies also shift with context. Annotation (AN) becomes more preferred in the Textbook setting, where additional explanation supports comprehension, but less preferred in the Movie setting, where brevity and fluency are more important. This aligns with the intuition that different translation settings impose different constraints on acceptable output.

Overall, these results highlight the importance of considering translation context and suggest that a single default translation strategy is insufficient across use cases.

\subsection{Strategy preference varies across source languages}

\begin{figure}
    \centering
    \includegraphics[width=\linewidth]{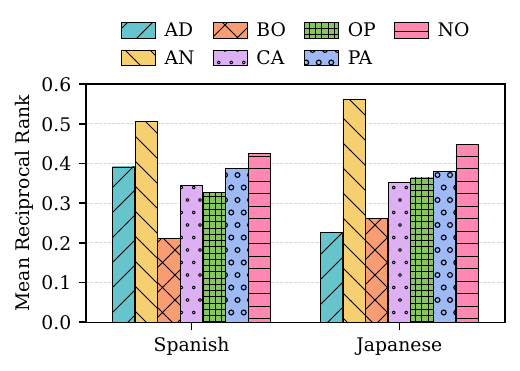}
    \caption{Mean Reciprocal Rank for each compensation strategy based on source language.  Adaptation is much more preferred by Spanish-speakers than Japanese-speakers.}
    \label{fig:lang}
\end{figure}

Figure \ref{fig:lang} compares preferences for Spanish and Japanese inputs.
Annotation (AN) is the most preferred strategy for both languages, but the preference is stronger for Japanese. This reflects greater linguistic and cultural distance between Japanese and English, increasing the need for explicit explanation.
In contrast, Adaptation (AD) performs relatively well for Spanish but is the least preferred strategy for Japanese. This suggests that adaptation is more difficult when structural and cultural differences between languages are larger, making it harder to preserve meaning through modification.

\subsection{Qualitative analysis}

Qualitative feedback from annotators provides additional insight into these trends.  Results often deviate from expectations; for example, Annotation (AN) remains highly ranked even in the Movie context, where shorter outputs are more practical.
One possible explanation is that annotators do not consistently incorporate contextual instructions into their judgments.  Some annotators do not reference context in their explanations of rankings.
Qualitative responses also help explain the low preference for Borrowing (BO), with annotators frequently noting that translations containing source-language words feel incomplete or incorrect.


\subsection{Application: Detecting Untranslatability}

As a simple demonstration of the dataset’s utility, we consider the task of detecting whether a given sentence exhibits untranslatability. Such a classifier could identify inputs that require specialized handling, such as selecting an appropriate compensation strategy or deferring to human translation.
Using a small evaluation setup, we find that incorporating examples from our dataset improves classification performance compared to a zero-shot baseline. This experiment illustrates how the structured representation of untranslatability provided by our framework can support downstream tasks. Details of this experiment are in Appendix~\ref{sec:class}.
\section{Discussion and Future Work}\label{sec:discuss}

Beyond the immediate findings of this work, we hope the proposed framework helps establish untranslatability as a primary problem in NLP rather than a collection of isolated translation edge cases. By organizing diverse phenomena under a shared ontology and linking them to explicit cStrats, the framework provides a common vocabulary for discussing translation under semantic mismatch.

Our results also suggest several directions for future MT research. 
Current systems typically optimize for a single default translation behavior, yet human preferences vary depending on the type of untranslatability and the translation setting. 
This motivates future systems to explicitly reason about when different translation strategies apply.
The dataset introduced in this work may support several downstream applications beyond those explored here. For example, it could be used to train systems that predict compensation strategies, generate context-aware translations, or identify inputs requiring specialized handling. Because the dataset also includes human preference rankings, it may additionally support preference-based training methods such as DPO \cite{rafailov2024directpreferenceoptimizationlanguage}.
More broadly, we hope this framework encourages future work that treats translation as a process of negotiating meaning across linguistic and cultural boundaries rather than solely optimizing semantic equivalence.
\section{Conclusion}\label{sec:conc}


We introduce the first structured framework for studying untranslatability in machine translation, defining an ontology of untranslatability types and a taxonomy of compensation strategies. By operationalizing this framework into a multilingual dataset, we enable systematic analysis of translation under semantic mismatch.
Our human preference study provides initial evidence that translation quality depends not only on fidelity, but also on the strategy used, with context-augmenting approaches often preferred over standard outputs. These findings suggest a gap between current MT objectives and human expectations and motivate viewing translation as a structured process involving both identifying mismatches and selecting appropriate strategies.
We view this work as a step toward strategy-aware machine translation and hope it supports future work on modeling and evaluating translation beyond one-to-one equivalence.

\newpage
\section*{Limitations}

We only work with Spanish and Japanese as source languages and English as target languages.  Results may differ with different source and target languages beyond the ones we investigate.  

We use GPT-4o as our LLM, which may have its own inherent biases.  We also observe systematic failure modes in generation. For example, the Pun (PN) type occasionally overfits to surface patterns in the prompt (e.g., punctuation or formatting cues), leading to less diverse outputs. These cases highlight challenges in controlling LLM-based generation for nuanced linguistic phenomena.  We include additional failure case analysis in Appendix \ref{sec:fail}.

Our human annotators are hired by Prolific, which may be biased in the specific annotators they allow to be matched with research projects.

\bibliography{custom}

\appendix

\section{uType and cStrat Details}
\label{sec:appendix-utype}

In this section we provide more robust definitions and examples for our ontology of untranslatability and taxonomy of compensation strategies.

\subsection{Linguistic Untranslatability}

\begin{table*}[]
    \centering
    \begin{tabularx}{\textwidth}{|c|X|X|X|}
    \hline
        Type & Example & Translation & Explanation  \\
        \hline
        NW & Tuvimos una sobremesa muy agradable. & We had a very nice sobremesa. & \emph{sobremesa} is a novel word in Spanish referring to the time spent chatting after a meal is had with peers. \\
        \hline
        SP & \begin{CJK*}{UTF8}{gbsn} 我叔叔很高。
        \end{CJK*} & My uncle is tall. & \begin{CJK*}{UTF8}{gbsn}
            叔叔
        \end{CJK*} (father's younger brother) has the underspecified translation `uncle' in English. \\
        \hline
        SX & \begin{CJK*}{UTF8}{gbsn} 这个电影我们都很喜欢。 \end{CJK*} & We all really like \emph{this movie}. & In Chinese, the object can be moved to the beginning of the sentence for emphasis.\\
        \hline
        VF & \begin{CJK*}{UTF8}{ipxm} 行きますか？うん、行く。\end{CJK*} & Are you going (formal)? Yeah, I'll go (informal). & Japanese verbs can be conjugated according to formality.\\
        \hline
        OI & Los profesores and las profesoras son diferentes. & The (male) teachers and the (female) teachers are different. & Many nouns referring to people in Spanish can be inflected to include gender in their meaning. \\
        \hline
        PH & \begin{CJK*}{UTF8}{gbsn} 施氏食狮史。\end{CJK*} & Shi's history of lion-eating & The Chinese characters all have the `shi' sound. \\
        \hline
        OR & \begin{CJK*}{UTF8}{ipxm}「wwww草」とコメントを残した。\end{CJK*} & They left a comment "I'm dead". & In Japanese, the character for `grass' means something like `lol' because `wwww', which also means something like `lol', looks like grass.\\
        \hline
        PN & Perdona, el pan está blando. Pues dile que se calle. & Excuse me, the bread is bland.  Well, tell it to shut up. & In Spanish, `bland' sounds like `talking'. \\
        \hline
        SL & \begin{CJK*}{UTF8}{gbsn} 我打游戏特别牛逼。\end{CJK*} & I’m absolutely badass at gaming. & \begin{CJK*}{UTF8}{gbsn} 牛逼 \end{CJK*} is a slang word related to a cow's anatomy, so it loses some of its literal meaning when translated. \\
        \hline
        ID & \begin{CJK*}{UTF8}{ipxm} 猿も木から落ちる。\end{CJK*} & Even the monkey falls from the tree. & This has a figurative meaning of `Everyone makes mistakes'.\\
        \hline
        RE & Se llama María como la madre de Jesús. & She's called Maria like the mother of Jesus. & In Spanish, Mother Mary's name is Maria.  These names would lose significance when translated to languages of non-Christian culture. \\
        \hline
        HI & \begin{CJK*}{UTF8}{ipxm} 古墳時代、飛鳥時代、奈良時代、平安時代。\end{CJK*} & Kofun period, Asuka period, Nara period, Heian period. & These eras can be translated literally, but they would not evoke the same connotations between people familiar and unfamiliar with Japanese history. \\
        \hline
        CU & Le di dos besos y me fui. & I gave her two kisses and left. & In Spain a common greeting and goodbye is to kiss the air while bumping cheeks together on both sides.\\
        \hline
    \end{tabularx}
    \caption{A single example for each uType.  Definitions of each Type can be found in Section \ref{sec:appendix-utype}. of the Appendix.}
    \label{tab:types}
\end{table*}

The two supertypes and their contained uTypes for the Linguistic domain are described below:

\paragraph{Vocabulary}  For Vocabulary, we consider word-level differences across languages.  We establish two specific types within Vocabulary: Novel Words (NW) and Specificity (SP).  Both types have to do with single words that exist in one language and not another, but the types differ in what makes those words untranslatable.  Novel Words include words that describe specific things, abstract or concrete, that are  not relevant to or do not exist in other cultures.  An example is `sobremesa' which is a Spanish word for the time spent chatting after a meal.  On the other hand, Specificity includes words that have a seemingly valid translation but are missing some extra nuance, making them over- or under-specified in reference to another language.  One example would be how Chinese has multiple words for `uncle' depending on specifics of the relationship.

\paragraph{Grammar} For Grammar, we consider  intra- and inter-word level linguistic differences.  For example, the Syntax type (SX) refers to how flexible word order changes in one language can affect the nuanced meaning while word order may be more rigid in another language.  For example, one can state the object first in Chinese for emphasis while it must come after the verb in English. On the intraword level, we have Verb Forms (VF), which refers to how many languages have ways to alter concepts like tense, mood, etc. within a verb through conjugation.  For example, the subjunctive mood in Spanish conveys hypotheticals or opinions.  Finally, Other Inflections (OI) covers other grammatical properties that affect nonverbs such as inflections on nouns to account for gender or number (like profesor vs. profesora for male and female teacher in Spanish).

\subsection{Figurative Untranslatability}

The five uTypes for the Figurative domain are described below:

Phonetics (PH) refers to the sounds within the utterance.  Alliteration or assonance would be examples of phonetic untranslatability.  A repetition of the same sound in multiple words in one language can create a specific effect for the listener, which would typically be impossible to maintain in a faithful translation.  An example would be `Peter Piper picked a peck of pickled-peppers.'

Orthography (OR) refers to the ``spelling'' or ``look'' of the words in a piece of writing.  Logographic scripts like Chinese or Hieroglyphs cause the bulk of this type of untranslatability.  However, even alphabet-based writing can have examples in this category, as each letter has a specific shape, which can have extra significance.  An example would be how `wwww' means `lol' in Japanese but since the w's look like grass, the character for grass, `\begin{CJK*}{UTF8}{gbsn} 草
        \end{CJK*}' can also mean `lol.'

Pun (PN) refers to when a word or phrase has multiple meanings in one language but not the other language.  Homophones and homonyms are common examples of this.  They are often used in comedy.  An example is `Is your refrigerator running? Better go catch it!'

Idiom (ID) refers to a combination of words which contains a literal meaning that has evolved over time in the language to also contain a figurative meaning.  Oftentimes, this requires two translations to fully capture the meaning in a different language (i.e. the literal and figurative translations). Occasionally, idioms have “perfect” translations across languages, which especially occurs when the languages are related and the idioms may have derived from an idiom in a common ancestral language.  An idiom in English is `to shoot yourself in the foot.'

Finally, Slang (SL) refers to words and phrases that are used in speaking more often than writing and particularly in informal contexts.  Often slang words are euphemisms or synonyms for other words that exist in the language used in a more general context.  An example for slang in English would be `He's got rizz.'

\subsection{Cultural Untranslatability
}

The uTypes for the Cultural domain are described below:

Religion (RE) refers to language which contains religious or philosophical significance and is therefore untranslatable to languages/cultures unfamiliar with that religion/philosophy.  Here, while the word or phrase may be literally translatable, the extra religious meaning would be lost in translation and is therefore untranslatable.  An example in English would be `Who will be the sacrificial lamb?'.

History (HI) refers to historical events within a specific culture which is not understood in other cultures.  Any reference to that historical event would be lost in translation.  An example in English is `Let's have a tea party in Boston'.

Finally, Customs (CU) is a more-encompassing category.  It refers to other societal significance that certain objects or concepts may contain within a specific culture.  Examples would be superstition, cultural perception of color, niceties etc.  For example, `This building has no 13th floor.'

In this section, we show tables for examples of the 13 uTypes and 6 cStrats.  See sections \ref{sec:ctype} and \ref{sec:cstrat} for definitions.

\subsection{Compensation Strategies}

\begin{table*}[]
    \centering
    \begin{tabularx}{\textwidth}{|c|X|X|X|}
    \hline
        Strategy & Example & Translation & Explanation  \\
        \hline
        AD & \begin{CJK*}{UTF8}{ipxm} うちの庭には二羽ニワトリがいる。 \end{CJK*} & She sells sea shells by the sea shore. & The original Japanese sentence is a tongue-twister about chickens in a yard, repeating the sounds `ni-wa'.  An Adaptation strategy would use an English tongue-twister. \\
        \hline
        AN & \begin{CJK*}{UTF8}{gbsn} 一朵花为什么好笑?因为它有梗。
        \end{CJK*} & Why is a flower funny? Because it has a stem. (Stem means joke in Chinese). & Here, the source sentence includes a pun (PN).  The Annotation strategy solves this by providing a literal translation with an aside that explains the extra meaning. \\
        \hline
        BO & \begin{CJK*}{UTF8}{ipxm}外国人は敬語を学ばないといけません。\end{CJK*}& Foreigners must learn keigo. & \begin{CJK*}{UTF8}{ipxm}敬語\end{CJK*} (keigo) refers to specific rules for inflections and word changes to express politeness and formality in Japanese. `Honorifics' may be a suitable translation though the Borrowing strategy would simply preserve the term as is. \\
        \hline
        CA & \begin{CJK*}{UTF8}{gbsn} 给他听爵士乐简直就是对牛弹琴。\end{CJK*} & Showing him jazz is like playing to a cow. & \begin{CJK*}{UTF8}{gbsn} 对牛弹琴\end{CJK*} (playing to a cow) is a Chinese idiom for offering something valuable to someone who cannot appreciate it.  The Calque Strategy translates the idiom literally and relies on the reader to extract the extra meaning through context or familiarity with the foreign idiom. \\
        \hline
        OP & Si tuviera más dinero, iría de vacaciones. & If I/they had more money, I/they would go on vacation. & There is a specificity mismatch in which the subject is unclear in Spanish as pronouns can be dropped.  English requires the pronoun to be specified, so the Options Strategy would offer multiple valid translations to account for multiple possible meanings. \\
        \hline
        PA & Uf, iba pedo anoche. & Oof, I was so drunk last night. & `Ir pedo' is a figurative slang Spanish phrase that describes being very drunk.  The Paraphrase Strategy works by simply providing an explanation of the meaning in plain language. \\
        \hline
     
    \end{tabularx}
    \caption{A single example for each cStrat.  Definitions of each cStrat can be found in Section \ref{sec:appendix-utype} of the Appendix.}
    \label{tab:strats}
\end{table*}

The six cStrats are described below:

Adaptation (AD) occurs when the translation replaces words and phrases in the source text/utterance in order to prioritize the untranslatable parts of the source over the literal meaning.  For example, a joke in one language about a certain object may have a literal translation, but the adaptation strategy would replace the joke with a joke that has a similar type of humor but may differ greatly in literal semantics.  Adaptation can occur on a smaller level as well, such as translating an object with a certain cultural significance into a completely different object with the same significance in the target culture.

Annotation (AN) occurs when the translation contains an aside or footnote appended to the official translation in order to fully capture the multiple layers of meaning in the source text/utterance.  Commonly, the official translation will be a literal translation with the figurative/cultural/nuance portion described in the note.

Borrowing (BO) occurs when the translation recognizes the untranslatability of the words or phrases and opts to simply use them in their source-language form in the target-language translation.  This is particularly common for the Novel Words category of untranslatability discussed above.  We often use borrowing when translating names as well.  

Calque (CA) is similar to borrowing and refers to taking an untranslatable word or phrase and translating it literally with the intent to recreate the same extra meaning it has in the source language.  For example, an idiom could be translated literally into a target language.  This would typically not make sense on the figurative level, but with repeated use over time, the literal translation would take on the original source’s figurative meaning. 

Options (OP) is a strategy for when there are specificity differences between the two languages in question.  For example, if one language specifies gender while another does not, a source sentence in the one that does not may require a translator to offer two different translations to account for the unknown gender.

Paraphrase (PA) refers to restating the original source with new and additional words in order to provide more clarity in the target.  With Paraphrase, literal translation is essentially abandoned and a representation of the extra meaning is provided.  For example, instead of replacing an idiom with one of similar figurative meaning (AD) or literally translating it (CA), a paraphrase would simply provide the relevant meaning of the idiom in non-figurative language.

\section{Dataset Details}\label{sec:details}

\begin{figure*}
    \centering
    \includegraphics[width=\linewidth]{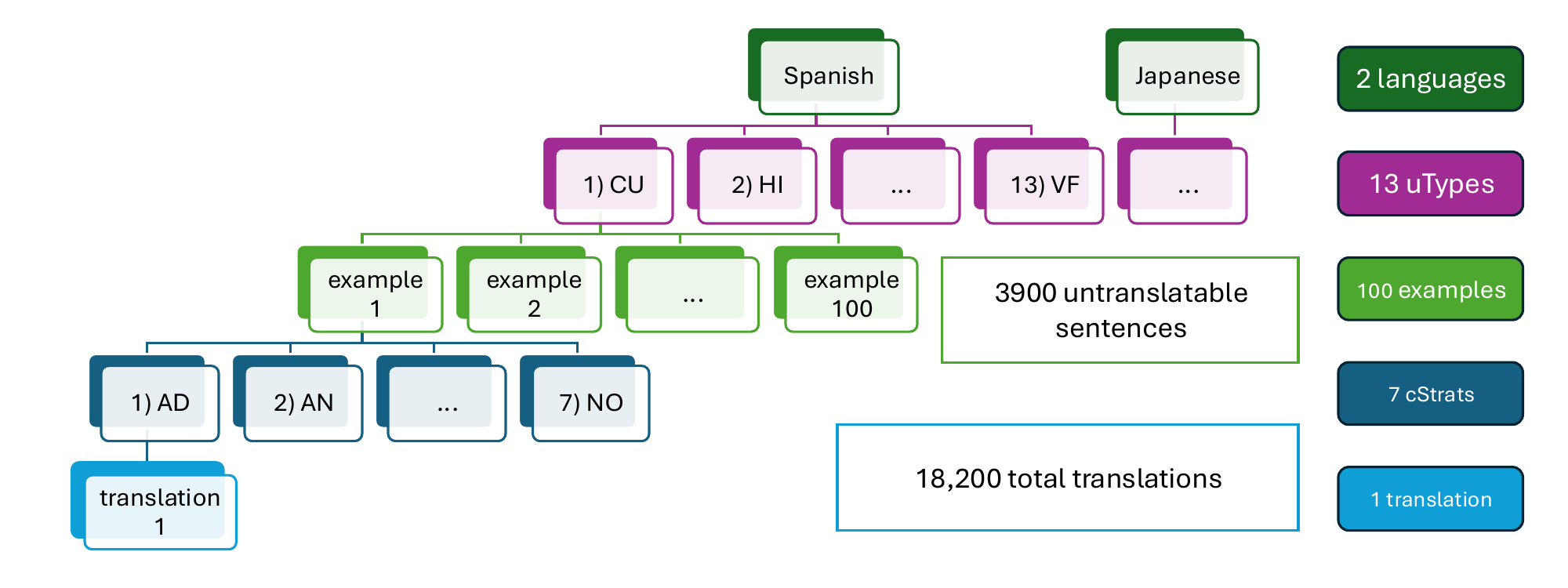}
    \caption{Visualization of the breakdown of the dataset.  We generated 18,200 English translations in total.}
    \label{fig:data}
\end{figure*}

A visualization of the breakdown of the dataset is shown in Figure \ref{fig:data}.  We originally planned to include Chinese, but due to unsatisfactory performance with GPT-4o and lack of language expertise from the first author, its inclusion will be delayed to a future release.

We also show the results from the data verification process in Tables \ref{tab:verif_es} and \ref{tab:verif_jp}. 130 examples were verified by language experts in each language, making up a representative 10\% of the dataset.

\begin{table}[]
    \centering
\begin{tabular}{|l|rrr|}
\hline
uType & \multicolumn{1}{l}{approved} & \multicolumn{1}{l}{checked} & \multicolumn{1}{l|}{percentage} \\
\hline
cu    & 10                            & 10                           & 1                              \\
hi    & 10                            & 10                           & 1                              \\
id    & 10                            & 10                           & 1                              \\
nw    & 9                            & 10                           & 0.9                           \\
oi    & 10                            & 10                           & 1                              \\
or    & 9                            & 10                           & 0.9                            \\
ph    & 10                            & 10                           & 1                              \\
pn    & 8                            & 10                           & 0.8                            \\
re    & 10                            & 10                           & 1                              \\
sl    & 9                            & 10                           & 0.9                            \\
sp    & 10                            & 10                           & 1                              \\
sx    & 10                            & 10                           & 1                              \\
vf    & 10                            & 10                           & 1                              \\
\hline
avg   & 9.615                  & 5                           & 0.9615 \\
\hline
\end{tabular}
    \caption{In-house language experts reviewed generated sentences of each uType to verify their belonging to the uType in question.  96\% of sentences were found to be valid examples in Spanish.}
    \label{tab:verif_es}
\end{table}

\begin{table}[]
    \centering
\begin{tabular}{|l|rrr|}
\hline
uType & \multicolumn{1}{l}{approved} & \multicolumn{1}{l}{checked} & \multicolumn{1}{l|}{percentage} \\
\hline
cu    & 10                            & 10                           & 1                              \\
hi    & 10                            & 10                           & 1                              \\
id    & 10                            & 10                           & 1                              \\
nw    & 8                            & 10                           & 0.8                          \\
oi    & 10                            & 10                           & 1                              \\
or    & 10                            & 10                           & 1                            \\
ph    & 9                            & 10                           & 0.9                              \\
pn    & 8                            & 10                           & 0.8                            \\
re    & 10                            & 10                           & 1                              \\
sl    & 10                            & 10                           & 1                            \\
sp    & 9                            & 10                           & 0.9                              \\
sx    & 10                            & 10                           & 1                              \\
vf    & 10                            & 10                           & 1                              \\
\hline
avg   & 9.538                  & 5                           & 0.9538 \\
\hline
\end{tabular}
    \caption{In-house language experts reviewed generated sentences of each uType to verify their belonging to the uType in question.  95\% of sentences were found to be valid examples in Japanese.}
    \label{tab:verif_jp}
\end{table}
\section{UI for Preference Annotation}

\begin{figure*}
    \centering
    \includegraphics[width=\linewidth]{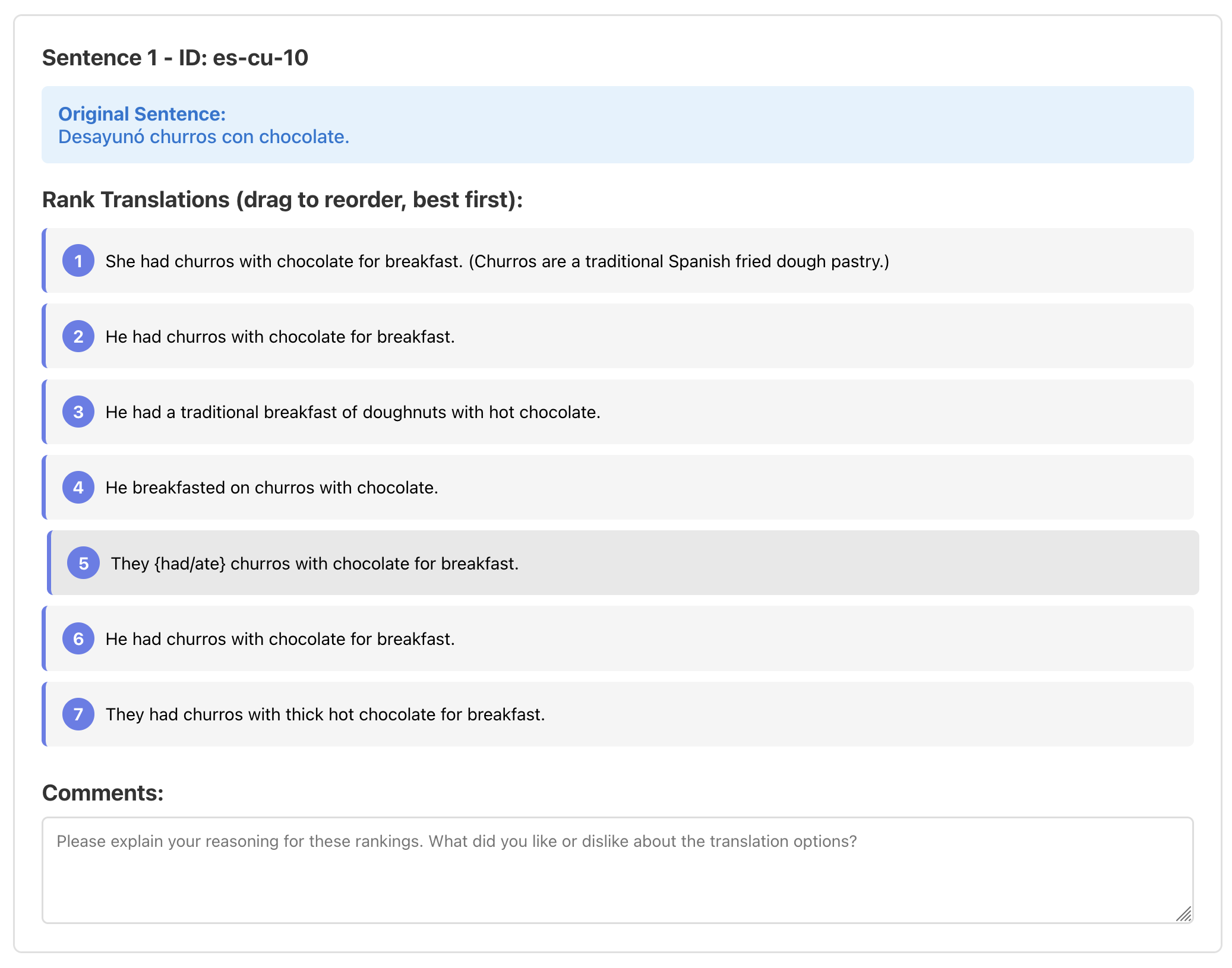}
    \caption{UI for Translation Preference Ranking task shown to bilingual annotators.}
    \label{fig:ui}
\end{figure*}

\begin{figure*}
    \centering
    \includegraphics[width=\linewidth]{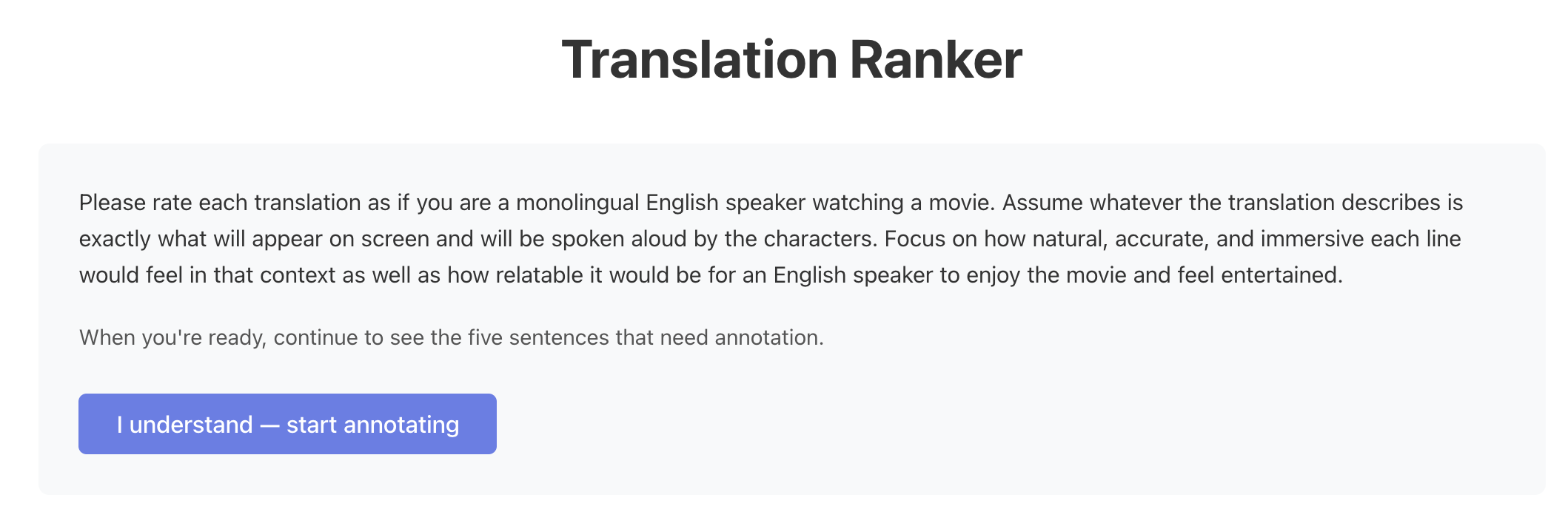}
    \caption{UI for context-based preference evaluation of translations.  Users are shown a paragraph that describes the translation setting before being asked to rank translations.}
    \label{fig:context-ui}
\end{figure*}

Here we show the UIs for the preference annotation collection from bilingual speakers.  They are shown in Figures \ref{fig:ui} and \ref{fig:context-ui}.

\section{Additional Results and Analysis from Human Studies}\label{sec:add}

\begin{figure*}
    \centering
    \includegraphics[width=\linewidth]{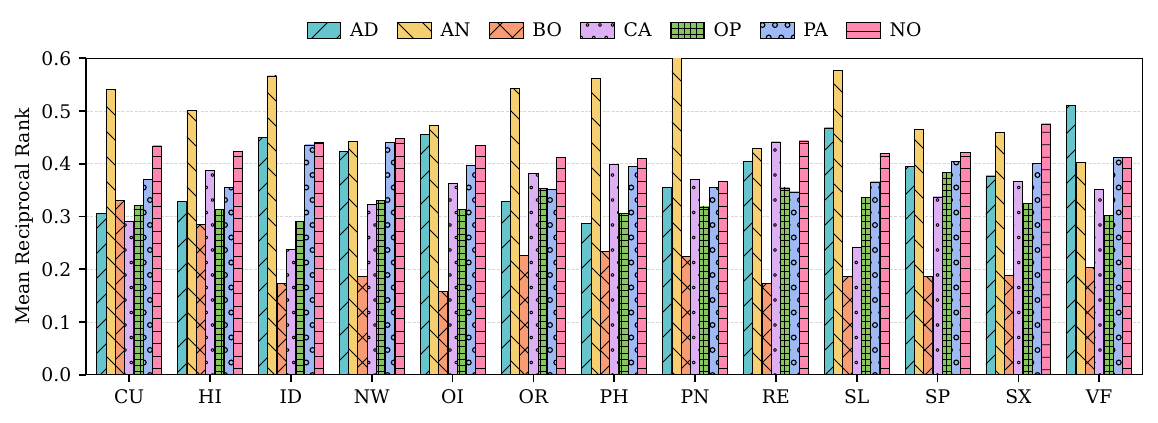}
    \caption{Results for General Context Spanish Preference Rankings.  Annotaion is the most preferred strategy on average and for almost all uTypes.  Different uTypes have different distributions of cStrat preference.}
    \label{fig:gen-es}
\end{figure*}

\begin{figure}
    \centering
    \includegraphics[width=\linewidth]{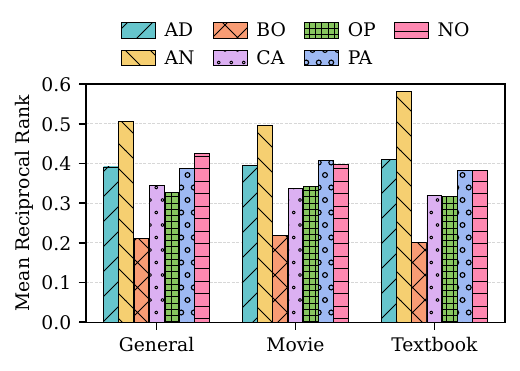}
    \caption{Mean Reciprocal Rank for each Compensation Strategy based on Translation Context.  We note similar trends with Annotation still winning out for all contexts and Borrowing consistently rated as the worst.}
    \label{fig:mmr-context}
\end{figure}

This section includes the figures for additional results and analysis of the human preference studies.  See Figure \ref{fig:gen-es} and Figure \ref{fig:mmr-context} for takeaways.

\section{Detecting Untranslatability}\label{sec:class}

We anticipate a variety of uses for our dataset.  To demonstrate explicitly one such use, we consider a classifier for detecting whether a given sentence is an instance of untranslatability or not.  This is useful as a solution to the untranslatability problem, as such a classifier could be used to triage `problem sentences' in large MT endeavors.  An MT model translates the sentences that are classified as translatable directly while those labeled as untranslatable are passed on to human translators.  This optimizes the balance between efficiency and quality as only the necessary sentences then undergo the more expensive human translation.

We create an untranslatability classification test set by generating 100 basic sentences in Spanish and combine them with the last 10 examples for each untranslatability type.  The full test set thus has 230 total sentences, 100 labeled as translatable and 130 as untranslatable.  We consider claude-opus-4-7\footnote{https://www.anthropic.com/news/claude-opus-4-7} with zero-shot prompting as a baseline and compare it to using 13 in-context examples of untranslatability from our dataset.  The results for accuracy of classification are shown in Table \ref{tab:class}.

\begin{table}[]
    \centering
    \begin{tabular}{|c|c|}
    \hline
        Model & Accuracy \\
        \hline
        Zero-shot & 0.704 \\
        With examples & \textbf{0.761} \\
        \hline
    \end{tabular}
    \caption{Accuracy for detecting untranslatable sentences.  The model that uses our untranslatability dataset for examples performs significantly better than a zero-shot model.  Our dataset is useful for detecting untranslatability to optimize between efficiency and quality in large-scale MT solutions.}
    \label{tab:class}
\end{table}

An even more robust solution is possible with our dataset by using the entirety of the untranslatable sentence generations and fine-tuning a language model to more accurately detect instances of untranslatability.  This along with many other applications make up possible uses of our dataset.  For brevity, we demonstrate this one solution using in-context examples to provide insight into the usefulness of the dataset.
\section{Failure Cases}\label{sec:fail}

As shown by the human verification studies in Tables \ref{tab:verif_es} and \ref{tab:verif_jp}, the generated dataset was 95\% valid.  This means there were some cases where the model failed to generate true examples of certain uTypes.  The frequency of these failures is small enough that it does not bring into question the utility of the dataset.  However, it does open the door toward failure cases analysis to consider how we may improve models to generate examples of different uTypes more reliably.  Categories we find issues in include NW, OR, PH, PN, SL and SP.

For NW, we find the model sometimes includes words that actually do have valid one-to-one translations in English.  These words are typically less common words such as `estribillo' which refers to the repeated part of a song (simply `chorus' in English).

OR shows issues in the Spanish section of the dataset.  In these failure cases, the sentence does not refer to specific letters and their shapes.  In one example, the generated sentences talks about `vocales' (vowels) but does not mention anything related to specific letters or their shapes.

PH shows issues in the Japanese section of the dataset.  In these cases, the sentence does not have any specific phonetic quality that makes it untranslatable.  In the failure cases, it does make some repetition but not enough that would signal an intentional phonetic pattern.

SL has some issues in the Spanish section of the dataset.  Sometimes the model generates phrases that are perhaps said in informal contexts but do not represent slang.  One example uses `bajar el perfil' which is to lower the profile or keep a low profile.  This could be said between friends but it has a very literal meaning without slang words.

Finally, we find some issues with SP in Japanese.  Sometimes the words in the generated example do not actually over- or under-specify when compared to English.  For example, \begin{CJK*}{UTF8}{ipxm}星がきれい。\end{CJK*} simply means the stars are pretty.

\end{document}